%% file: 00-main.tex
\documentclass[11pt]{article}
\usepackage{coling2020}
\usepackage{times}
\usepackage{newtxmath}
\usepackage{url}
\usepackage{latexsym}
\usepackage{xspace}
\usepackage{comment}
\usepackage{graphicx}
\usepackage[disable,textsize=footnotesize, color=green]{todonotes}
\definecolor{Gray}{gray}{0.95}
\newcommand{\cntodo}[1]{\todo[color=yellow]{#1}}

\newcommand{\jttodo}[1]{\todo[color=pink]{#1}}

\newcommand{\corpus}{\texttt{GAQCorpus}\xspace} 
\newcommand{\corpuslong}{Grammarly Argument Quality Corpus\xspace} 

\newcommand{\tvsp}{{\sc TvsP}\xspace}
\usepackage{subcaption}
\usepackage{booktabs}
\usepackage{makecell}
\usepackage{tabularx}
\usepackage{colortbl}
\usepackage{wrapfig}
\usepackage{pdfpages}
\usepackage{multirow}
\usepackage{enumitem}

\colingfinalcopy 

\title{Adapting Argument Quality Annotation to Non-Standard Domains}
\title{\corpus: A Domain-diverse Corpus \\ for  Theory-based Argument Quality Assessment}
\title{Creating a Domain-diverse Corpus\\ for Theory-based Argument Quality Assessment}
\author{Lily Ng\textsuperscript{1}\thanks{~~Equal contribution.}, Anne Lauscher\textsuperscript{2}\footnotemark[1], Joel Tetreault\textsuperscript{3}, Courtney Napoles\textsuperscript{1} \vspace{0.3em} \\
  \textsuperscript{1}Grammarly \\
  \textsuperscript{2}Data and Web Science Group, University of Mannheim, Germany \\

  \textsuperscript{3}Dataminr, Inc.
  \\ 
  
  \textsuperscript{1}{\tt first.last@grammarly.com}, 
  \textsuperscript{2}{\tt anne@informatik.uni-mannheim.de}, \\
  \textsuperscript{3}{\tt jtetreault@dataminr.com}
}

\date{}
\begin{document}
\maketitle

\begin{abstract}
\jttodo{make sure to revisit the title: what if we just called it the name of the corpus}
Computational models of argument quality (AQ) have focused primarily on assessing the overall quality or just one specific characteristic of an argument, such as its \textit{convincingness} or its \textit{clarity}. 
However, previous work has claimed that assessment based on theoretical dimensions of argumentation could benefit writers, but developing such models has been limited by the lack of annotated data. 
In this work, we describe \corpus, the first large, domain-diverse annotated corpus of theory-based AQ. We discuss how we designed the annotation task to reliably collect a large number of judgments with crowdsourcing, formulating theory-based guidelines that helped make subjective judgments of AQ more objective. 
We demonstrate how to identify arguments and adapt the annotation task for three diverse domains.    
Our work will inform research on theory-based argumentation annotation and enable the creation of more diverse corpora to support computational AQ assessment. 
\end{abstract}

\section{Introduction}
\input{1-introduction}

\section{Related Work}
\label{sec:related-work}
\input{2-related-work}


\section{Annotation Study}
In this section, we detail how we developed and designed our annotation task to enable efficient, reliable collection of theory-based AQ judgments with crowdsourcing. We validate \newcite{wachsmuth-etal-2017-argumentation}'s hypothesis that crowdsourced annotation of theory-based AQ is possible if the task is simplified.

\label{sec:annotation}
\subsection{Simplifying the task}\label{sec:simplify}
\input{3.2-simplifying-task}

\section{Data Domains}
\label{sec:domains}
\input{4-data-domains}

\section{A Theory-based AQ Corpus}
Applying the annotation task design and data selection described above, we created \corpus, containing 5,285 arguments across three domains, annotated for theory-based dimensions. All arguments were limited to have a length between 70 and 200 characters. 
Ratings were provided by the two groups of annotators described above, Experts (\S\ref{sec:simplify}) and the Crowd (\S\ref{sec:validating}). 
Each group judged 3,000 arguments, with about 1,000 arguments annotated by both groups for comparison. The size of the corpus is described in Table~\ref{tab:corpus-size}. Annotators worked with the domains in the following order: Debate forums, CQA forums, and Review forums. Before switching to a new domain, annotators completed a small study for calibration. 
All data and guidelines are available from \url{https://github.com/grammarly/gaqcorpus}.

\label{sec:analysis}
\input{5.1-quantitative-analysis}
\subsection{Qualitative Analysis}\label{sec:qualitative}
\input{5.2-qualitative-analysis}

\section{Conclusion}
\label{sec:conclusion}
\input{6-conclusion}

\section*{Acknowledgements}
The work of Anne Lauscher is supported by the Eliteprogramm of the Baden-Württemberg Stiftung (AGREE grant).
We thank our linguistic expert annotators for providing interesting insights and discussions as well as the anonymous reviewers for their helpful comments. We also thank Henning Wachsmuth for consulting us w.r.t. his previous work and Yahoo! for granting us access to their data.
\bibliographystyle{coling}
\bibliography{coling2020}

\end{document}


\title{Supplemental Material}
\maketitle
\section{Guidelines}
\subsection*{Annotation guidelines}

\subsubsection{Debate Forums}\includepdf[pages=-]{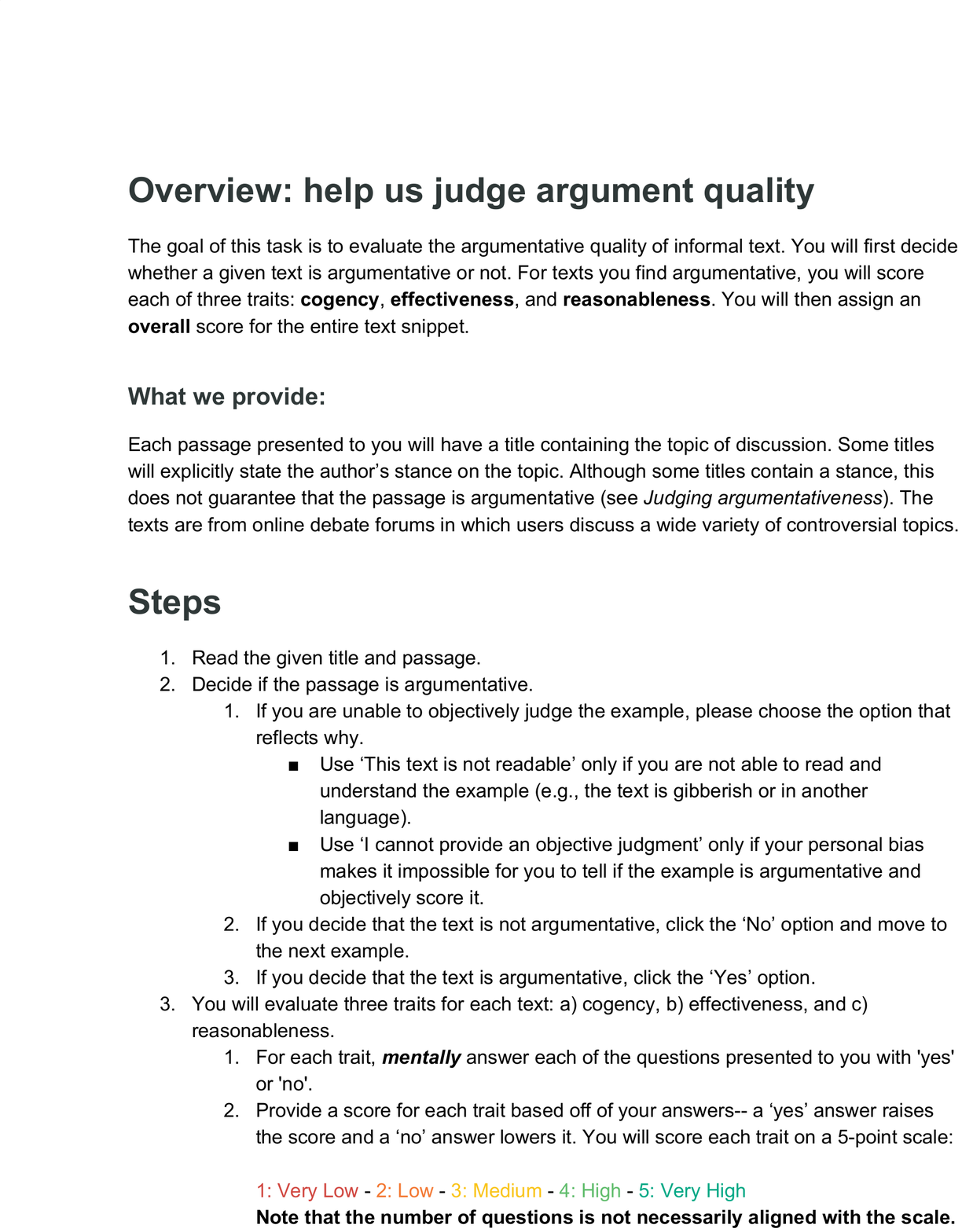}
\subsubsection{Q\&A Forums}\includepdf[pages=-]{debate_yahoo.pdf}
\subsubsection{Review Forums}\includepdf[pages=-]{yelp.pdf}

\input{7-appendix}

%% file: 1-introduction.tex
The notion of \emph{Argumentation Quality (AQ)} plays an important role in many existing argument-related downstream applications, such as argumentative writing support \cite{stab-gurevych-2017-recognizing}, automatic essay grading \cite{persing-ng-2013-clarity}, and debate systems \cite{toledo2019automatic}. For some of these applications, the idea is to automatically give feedback to users to help them improve their writing skills or assess their writing capabilities. For others, assessing AQ is an important step in a more complex pipeline for retrieving high-quality arguments. 

While grading overall AQ \cite{toledo2019automatic} or a specific conceptualization of AQ, such as \emph{prompt adherence} \cite{persing-ng-2014-prompt} is relatively well explored, researchers have noted the lack of work in so-called \emph{theory-based AQ}\footnote{In the following, we adopt the term ``theory-based AQ," which was proposed by \newcite{wachsmuth-etal-2017-computational} to indicate that the conception of AQ is specifically grounded in argumentation theoretic literature (and not in CL or NLP).} \cite{wachsmuth-etal-2017-computational}, which can be represented with a taxonomy characterizing overall AQ into several subdimensions and aspects, for instance, as \emph{logic} and \emph{rhetoric}, which therefore provides a more informative and targeted perspective. However, this holistic approach comes with the downside of higher complexity, especially when it comes to annotating textual corpora, which are required for training and developing common computational approaches (see, e.g., \newcite{gretz2019large}).  In a small study, \newcite{wachsmuth-etal-2017-argumentation} demonstrate that theory-based AQ annotations can be done both by trained experts and by crowd annotators, though the authors acknowledge the high complexity and subjectivity of the problem and call for the simplification of theory-based AQ annotation in order to reliably create larger-scale corpora. To date, no work has tackled this challenge and accordingly, no larger-scale and no domain-diverse corpus of this kind exists. 
We aim to close this gap by describing our efforts to create \corpuslong (\corpus) \cite{lauscher2020rhetoric}, the largest and the only domain-diverse corpus consisting of 5,285 English arguments annotated with theory-based AQ scores across four dimensions.

Building on \newcite{wachsmuth-etal-2017-argumentation}, in this work, we modify the annotation task to be suitable for both experts and the crowd while preserving the theoretical basis of the taxonomy. We collect and annotate argumentative texts from web debate forums, 
as well as community questions and answers forums (CQA),\jttodo{is this referring to the Yahoo Answers domain?  If so, we should call it CQA - Community Questions and Answers. CN: TO DO in this and the COLING submission} 
and review forum texts, which are still understudied in computational AQ. The latter domains can consist of rather non-canonical arguments in that they exhibit a lack of explicitness of certain argumentative components;  are topic-wise more subjective; or consist of longer, more convoluted text. This makes assessing the quality of such arguments even more challenging, but downstream can result in a more robust model of computational AQ.

Given all these challenges, we work closely with trained linguists to adapt the annotation task, iterating over how best to approach these novel domains and simplify the annotation guidelines for crowdsourcing, allowing us to  collect a large number of judgments efficiently. 
We hope that our work fuels further research on theory-based computational AQ. Our approach to building \corpus can inspire and inform AQ annotation in new domains, enriching the domain-diversity of linguistic resources available in this space and consequently expanding computational approaches to AQ. 

\paragraph{Structure.} We start by surveying previous AQ annotation studies (\S\ref{sec:related-work}). Next, we describe our efforts to adapt and simplify the annotation task (\S\ref{sec:annotation}), which is followed by a discussion of the data domains (\S\ref{sec:domains}). \S\ref{sec:analysis} presents an analysis of the resulting corpus. Finally, we conclude our work and provide directions for future research (\S\ref{sec:conclusion}).

%% file: 2-related-work.tex
%

\begin{wrapfigure}{r}{0.48\textwidth}
  \centering
    \vspace{-15mm}
    \includegraphics[width=\linewidth,trim=0.3cm 0cm 0.3cm 0cm]{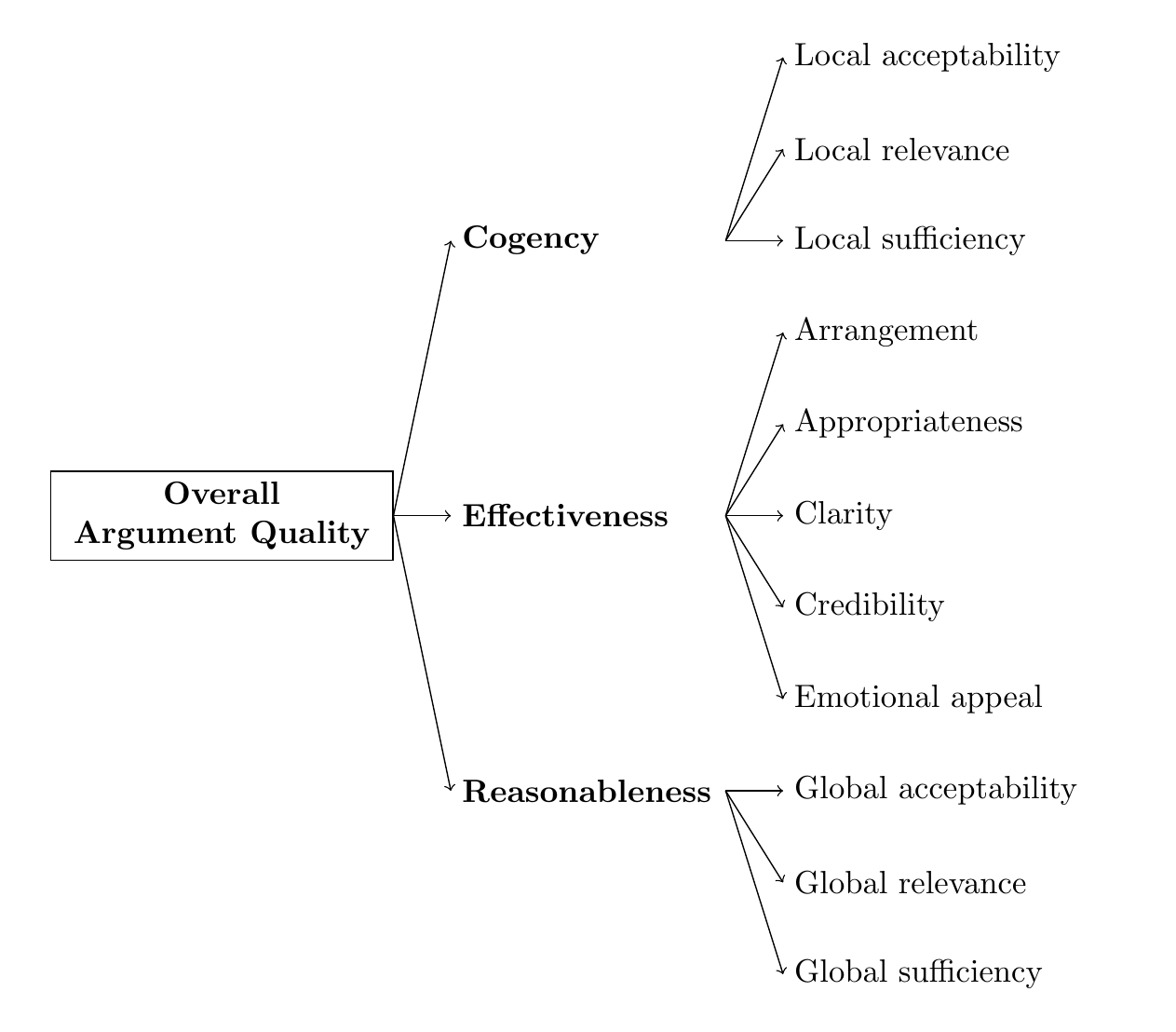}
  \caption{The taxonomy of theory-based argument quality aspects~\cite{wachsmuth-etal-2017-computational}.}
  \label{fig:tax}
\end{wrapfigure}
Most argumentation annotation studies have been conducted on student essays or web debates. Student essays have been annotated for thesis clarity \cite{persing-ng-2013-clarity}, organization \cite{Persingorganization}, and prompt adherence \cite{persing-ng-2014-prompt}, and \newcite{persing2015modeling} model argument strength rated on a 4-point Likert scale. Similarly,
\newcite{stab2016recognizing} annotate the absence of opposing arguments and \newcite{stab-gurevych-2017-recognizing} predict insufficient premise support in arguments. 
For web debates, \newcite{habernal-gurevych-2016-argument} conduct an annotation study in which they present debate arguments pairwise to crowd annotators, who then can choose the more convincing argument. 
\newcite{Persing:2017:WCY:3171837.3171856} also annotate the reasons why an argument receives a low persuasive power score. 

\newcite{wachsmuth-etal-2017-computational} developed a taxonomy of AQ synthesized from traditional works in argumentation theory, such as \newcite{kennedy2007aristotle}.
The full taxonomy is depicted in Figure \ref{fig:tax}, and defines the Overall AQ to consist of the following three subdimensions, each of which is itself defined by several finer-grained AQ aspects:

\smallskip\noindent \textbf{(1) Cogency} relates to the logical aspects of AQ, for instance, whether the an argument's premises are acceptable (local acceptability) or whether they can be seen as relevant for the conclusion (local relevance).
 
\smallskip \noindent \textbf{(2) Effectiveness} indicates the rhetorical aspects of an argument. Aspects of effectiveness include, for instance, its clarity or its emotional appeal.
 
\smallskip \noindent \textbf{(3) Reasonableness} reflects the quality of an argument in the overall context of the discussion, as, for instance, its relevance towards arriving at a resolution of the issue (global relevance).

\bigskip
\newcite{wachsmuth-etal-2017-argumentation} conducted a study in which crowd workers annotated 304 arguments for all 15 quality dimensions  (Figure~\ref{fig:tax}), 
and demonstrated that the theory-based and practical AQ assessments match to a large extent. 
Their findings indicate that theory-based annotations can be crowdsourced and that theory-based approaches can inform the practical view, especially. Most importantly, the authors conclude that the annotation task should be simplified to guarantee a reliable crowd-annotation process. 

Most recently, \newcite{toledo2019automatic} and \newcite{gretz2019large} crowdsourced overall argument quality by presenting pairwise arguments to annotators, who then had to select the argument ``they would recommend a friend to use that argument as is in
a speech supporting/contesting the topic." This is an extreme simplification of the task, which does not seem to lead to better agreement: the authors \cite{gretz2019large} report an average inter-annotator agreement of $\kappa=0.12$  and attribute the low score to the high subjectivity of the task. 
The authors conducted a theory-based annotation study in the spirit of \newcite{wachsmuth-etal-2017-computational} on a subset of the data (100 arguments) which indicated the highest correlation of the annotations with the effectiveness dimension. Later on, \newcite{lauscher2020rhetoric} 
empirically confirmed this observation using computational model predictions across the whole corpus.  


Building on this large body of work, we aim to facilitate the annotation of theory-based AQ in diverse domains of real-world argumentative writing and compare expert vs.~crowd annotations. Our study results in the largest English corpus annotated with theory-based argumentative quality scores.

%% file: 3.2-simplifying-task.tex

Before collecting any crowdsourced annotations, we conducted 14 pilot experiments with a group of four ``expert'' annotators, simplifying the \tvsp task design through their feedback and observations, as they provided both a deep understanding of the argumentation theory and practical experience annotating the arguments.
Each expert annotator was a fluent or native English speaker with an advanced degree in linguistics.
Experts underwent training, which included studying guidelines and participating in calibration tasks to analyze debate arguments from three sources: 
Dagstuhl-ArgQuality-Corpus-V2\footnote{\url{http://argumentation.bplaced.net/arguana/data}}, originally from UKPConvArgRank \cite{habernal-gurevych-2016-argument}; the Internet Argument corpus V2\footnote{\url{https://nlds.soe.ucsc.edu/iac2}} (IAC) \cite{abbott-etal-2016-internet}; and ChangeMyView,\footnote{\url{https://www.reddit.com/r/changemyview/}} a Reddit forum.
Through the pilots and subsequent debriefs with the experts, we made the following modifications to the annotation task of \newcite{wachsmuth-etal-2017-argumentation}:

\paragraph{(1) Reduce taxonomy complexity.} 
While \tvsp defined the task to score all $11$ AQ subaspects (Local Acceptability, Local Relevance, etc.), $3$ dimensions (Cogency, Effectiveness, Reasonableness), and overall AQ, we reduced the number of qualities scored by only  focusing on the $3$ higher-level dimensions plus overall AQ. As a result, annotators assessed an argumentative text in terms of $4$ scores instead of $15$ scores, and instead of $3$ different AQ levels, the simplified taxonomy is reduced to $2$. 

\paragraph{(2) Instruction Modifications.} We reworded the \tvsp dimension descriptions and added several examples to make the guidelines more understandable. 
As the annotators were not rating the $11$ AQ subaspects, we experimented with different methods to incorporate the subaspects into the guidelines. 
Instead of explaining the subdimensions in the guidelines and trusting crowd annotators to bear them in mind, we represented each subdimension as a yes/no question in the annotation task itself (Table~\ref{tab:subdimension_questions}). Our pilot experiments showed that presenting the questions without asking for a response eased the perceived complexity of the task while not affecting agreement.
\begin{table}[]
    \centering
    \small
    \begin{tabular}{llp{10cm}}
\toprule
Dimension & Subdimension & Question\\
\midrule
Cogency & Local Acceptability & Are the justifications for the argument acceptable/believable?\\
\arrayrulecolor{lightgray} \cmidrule{2-3}
 & Local Relevance & Are the justifications relevant to the author’s point?\\
 \cmidrule{2-3}
 & Local Sufficiency & Do the justifications provide enough support to draw a conclusion?\\
\arrayrulecolor{black}
\midrule
Effectiveness & Credibility & Is the author qualified to be making the argument?\\
\arrayrulecolor{lightgray} \cmidrule{2-3}
& Emotional Appeal & Does the argument evoke emotions that make the audience more likely to agree with the author?\\
\cmidrule{2-3}
& Clarity & Does the author’s language make it easy for you to understand what they are arguing for or against?\\
\cmidrule{2-3}
& Appropriateness & Is the author’s argument and delivery appropriate for an online forum?\\
\cmidrule{2-3}
& Arrangement & Did the author present their argument in an order that makes sense?\\
\arrayrulecolor{black}
\midrule
Reasonableness & Global Acceptability & Would the target audience accept the argument and the way it is stated?\\
 \arrayrulecolor{lightgray} \cmidrule{2-3}
 & Global Relevance & Does the argument contribute to the resolution of the given issue?\\
 \cmidrule{2-3}
 & Global Sufficiency & Does the argument address and adequately rebut counterarguments?\\
\arrayrulecolor{black}\bottomrule    
\end{tabular}
    \caption{Subdimensions represented as questions in the annotation task of debates.}
    \label{tab:subdimension_questions}
\end{table}

\paragraph{(3) Five-point scale.} While \tvsp collected judgments with a three-point rating scale (low, medium, high), we employ a five-point scale (very low, low, medium, high, very high, plus \textit{cannot judge}) to allow for more nuanced judgments, as the expert annotators found too great of a distance between the items on a three-point scale. Scales with 5--9 items have been shown to be optimal, balancing the informational needs of the researcher and the capacity of the raters~\cite{cox1980}. 
We experimented with both three- and five-point scales and found that the larger scale did not negatively affect inter-annotator agreement.

\subsection{Validating the Task Design}\label{sec:validating} 
Our finalized task design is as follows.  First, annotators decide whether a text is argumentative. If \textit{yes}, the three high-level dimensions are scored on a five-point scale and subaspect questions are presented to guide the annotator's judgment.  The Overall AQ is scored last, also on a five-point scale. 

\begin{wraptable}{r}{.55\linewidth}
\centering
{\small
\begin{tabular}{ l c c c c}
\toprule
& \multicolumn{1}{c}{Cogency} & \multicolumn{1}{c}{Effectiveness} & \multicolumn{1}{c}{Reasonableness} & \multicolumn{1}{c}{Overall} \\
\midrule
Ours & \textbf{0.46} 
& \textbf{0.48} 
& \textbf{0.48} 
& \textbf{0.55} \\
TvsP & 0.27 
& 0.38 
& 0.13 
& 0.43 
\\
\bottomrule
\end{tabular}
}
\caption{Agreement Dagstuhl ``gold" annotations and our crowdsourced annotations (Ours) compared to \tvsp.}
\label{tbl:dagstuhl}
\end{wraptable} 

\vspace{2mm} Before collecting annotations from the crowd, we validated our modifications subjectively and objectively. First, we ran a series of pilot tasks with our expert annotators. 
They initially annotated using the \tvsp guidelines and next worked with the simplified taxonomy. In follow-up discussions, the experts confirmed that the new task design reduced the time and cognitive load necessary to rate arguments, and that the guidelines were more understandable. These modifications make the task more approachable, which is vital when presenting it to (untrained) crowd-workers for larger-scale annotation. 

We validated the simplifications quantitatively by reproducing the study of \tvsp, which compared their crowd and ``expert'' annotations. To this end, we randomly sampled $200$ arguments from Dagstuhl-ArgQuality-Corpus-V2, which come with author-annotated ``gold'' ratings. 
We collected ratings from a crowd (10 ratings per item), following our simplified design\footnote{The only difference is that we used a 3-point scale to more fairly compare to the gold.} (\S\ref{sec:simplify}). 
All crowd contributors were native or fluent English speakers engaged through Appen (formerly Figure Eight).
Crowd contributors did not participate in calibration meetings and all feedback was relayed to contributors through a liaison.

We average the crowd ratings to obtain a single score for each argument and computed the inter-annotator agreement (IAA) with the ``gold'' annotations using  Krippendorff's $\alpha$ \cite{krippendorff2011computing} (Table~\ref{tbl:dagstuhl}). 
Even though the annotation scores are not strong, the IAA between our crowd annotators and the gold annotations generally
 surpasses the agreement scores reported by \tvsp. This is a highly nuanced and subjective task, which is reflected in the agreement levels.
 Based on these results and annotator observations, we conclude that our task guidelines and design allow for better (or at least comparable) quality crowdsourcing of theory-based AQ annotations. 

%% file: 4-data-domains.tex
In this work, we consider
three domains: \textit{Debate} forums, \textit{CQA} forums, and \textit{Review} forums.
While Debates are generally well-explored in computational AQ, 
we are unaware of any work involving CQA and Reviews. 
For each of these domains, we first identified items likely to be argumentative and then adjusted the guidelines in consultation with expert annotators, as described below. 



\paragraph{Debate forums.}
Of these three domains, \textit{Debates} is the most straightforward to annotate. Given a topic or motion, users can define their stance (\emph{pro}/\emph{contra}) and write an argument which supports it.
We included data from two online debate forums. ConvinceMe (CM) is a subset of the IAC, where users share their \emph{Stance} on a topic and discuss their point of view, with replies aiming to change the view of the original poster. Change My View (CMV) is a Reddit forum in which participants post their opinion on a topic and ask others to post replies to change their mind.
We sampled original posts from CMV, skipping any moderator posts, and the first reply to an original post from CM, in order to limit the context that annotators must consider when evaluating arguments. 
CMV posts always include the author's perspective in the title, while CM posts may or may not include a stance in the title. 
In the guidelines, we instruct annotators to judge a post by how successfully it justifies the author's claim. 

\paragraph{CQA.}
 In community questions and answers forums, users post questions or ask for advice, which other users can address. 
We experimented with arguments from Yahoo! Answers\footnote{\url{https://answers.yahoo.com/}} (YA).
When posting a question, a user can provide background information for their question (\textit{context}) and can later indicate which response is the \textit{best answer} to their question.
The forum's looser structure provides for a wide variety of content, which is appealing as a potential source of non-standard arguments, but challenging as many of the posts do not contain any arguments.
Through manual analysis, we identified three categories that frequently contained controversial topics, hypothesizing they would have a higher incidence of debates: \textit{Social Science $>$ Sociology}, \textit{Society \& Culture $>$ Other}, and \textit{Politics \& Government $>$ Law \& Ethics}. We empirically selected the category with the highest proportion of arguments in a study on Amazon Mechanical Turk (MTurk).
Qualified annotators\footnote{HIT approval rate $>=$ 97; HITs approved $>$ 500; Location = US} decided if question and best-answer pairs were argumentative. We collected 10 judgments for 100 pairs from each category and
aggregated judgments with a simple majority. \textit{Law \& Ethics} had the most argumentative posts (70\%, compared to \textit{Sociology} with 40\% and \textit{Society \& Culture} with 34\%), so we sampled posts from this category to annotate.

In the guidelines for this domain, we asked annotators to judge the argumentative strength of an answer with respect to how well it addressed the given question.
The guidelines and subdimension questions were altered to encourage this. 
One obstacle in pilot studies with expert annotators was posts offering, as many users solicited legal advice in the Law \& Ethics forum.
We decided to consider advice as argumentative as long as the author supported the advice with justification, which mirrors our general approach to the Argumentative dimension.

\paragraph{Reviews.}
 The third domain consists of restaurant reviews from the  Yelp-Challenge-Dataset\footnote{\url{https://www.yelp.com/dataset}}.
On Yelp, users write reviews of businesses and rate the quality of their experience from 1 (low) to 5 (high) stars.
Unlike the Debate and Q\&A forums, the format of Yelp does not support dialogue between users (i.e., users cannot directly reply to other users or posts), and so it is possible to present each post in isolation as a self-contained argument.
As most posts do not explicitly state a claim, we pose the star rating as a claim the user is making about the business, and the review as the argument supporting it.

Yelp reviews can be highly subjective in that each review is based on a single user's experience.
For instance, a user may rate a restaurant as 5-stars and write only \emph{The food was delicious} in their review.
To address this subjectivity, we asked annotators to judge the argumentative quality of each review with respect to how well it supported the rating provided.
Another challenge was defining what constituted a counterargument, as these have a very different character than counterarguments in debates (for example, \emph{Everyone says that the pizza crust is too thin here but that’s authentic!}).
In consultation with our experts, we defined counterarguments by the following characteristics: 1) addressing and rebutting the viewpoints of other reviews, 2) addressing and rebutting points that discredit the author’s rating, and 3) bringing up favorable points in an unfavorable review and vice versa.

\bigskip
\noindent Experts completed a series of pilots before each domain was presented to the crowd, using the task design described in \S\ref{sec:simplify}.
Expert agreement on novel domains (YA and Yelp) are shown in Table \ref{tab:pilots_domains}.
Feedback on the task and guidelines was gathered during calibration meetings with experts, and they were iteratively altered to be more clear and specific. 

\begin{wraptable}{r}{.6\linewidth}\vspace{-1.2cm}
\centering{\footnotesize
\begin{tabular}{lrrrrrp{5pt}c}\toprule
&Crowd &\multicolumn{3}{c}{Experts} &Overlap \\
\cmidrule(lr){2-2} \cmidrule(lr){3-5} \cmidrule(lr){6-6}
\# Annotators &10 &1 &2 &3 & 11--13&\hspace{8em}&\textbf{Total size}\\
\midrule
CQA &1,334 &626 &-- & 625 &500 &&\textbf{2,085}\\
Debates &1,438 &600 &-- &600 &538 &&\textbf{2,100}\\
Reviews &600 &200 &400 &-- &100 &&\textbf{1,100}\\
\bottomrule
\end{tabular}}\caption{Number of arguments annotated by experts and the crowd and the number of overlapping instances (annotated by both experts and the crowd) by domain.}
\label{tab:corpus-size}
\end{wraptable}


%% file: 5.1-quantitative-analysis.tex
\setlength{\tabcolsep}{3pt}
\begin{table*}[!t]
\parbox[t]{.48\linewidth}{

\strut\vspace*{-\baselineskip}\newline
\centering
{\footnotesize
\begin{tabular}{lcccc}
\toprule
Domain & Cogency & Effectiveness & Reasonableness & Overall\\
\midrule
CQA & 0.16 & 0.31 & 0.36 & 0.29\\
Debates & 0.22	& 0.33 & 0.20 & 0.33\\
Reviews & 0.41 & 0.19 & 0.21 & 0.34\\
\bottomrule
\end{tabular}}
\caption{\label{tab:pilots_domains} Agreement (Krippendorff's $\alpha$) between experts on pilot studies for CQA, Debates, and Reviews (146, 150, and 50 arguments, respectively).}
%
}
\hfill
\parbox[t]{.48\linewidth}{
\strut\vspace*{-\baselineskip}\newline
\centering
\small
\begin{tabular}{lcccc}\toprule
& Cogency &Effectiveness &Reasonableness &Overall \\\midrule
CQA  &0.42 &0.52 &0.52 &0.53 \\
Debates\hspace{-1em} &0.14 &0.11 &0.21 &0.19 \\
Reviews\hspace{-1em} &0.32 &0.32 &0.31 &0.33 \\
\bottomrule
\end{tabular}
\caption{IAA between the mean expert and crowd scores for Cogency, Effectiveness, Reasonableness, and Overall AQ.}\label{tab:agreement_expert_crowd}
}

\end{table*}

\subsection{Inter-annotator Agreements (IAA)\label{iaa}} 

\begin{wrapfigure}{r}{.54\linewidth}
    \centering
\vspace{-12mm}\includegraphics[width=\linewidth,trim=0.25cm 0.0cm 1.5cm 0cm]{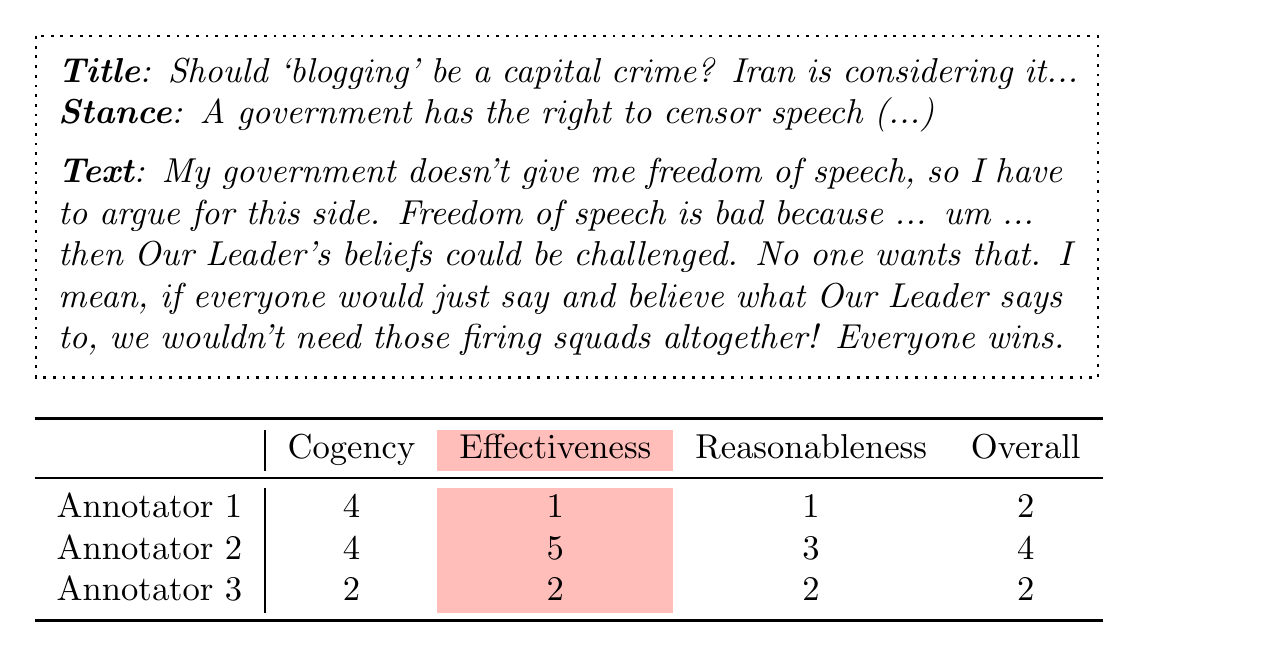}
    \caption{Example argument exhibiting disagreement in the Effectiveness dimension.}
    \label{fig:ex1}
\end{wrapfigure}

We assessed the quality of the crowd annotations by calculating the agreement between the experts and crowd workers on the overlapping portions of \corpus using the mean scores (Table~\ref{tab:agreement_expert_crowd}). 

For debate forums, the agreement is weak with  $\alpha \leq 0.21$, while for the CQA forums, the agreement is higher: $0.42$--$0.53$. These results suggest that the difficulty of the task is highly dependent on the domain. While our Debates data and the DS data both consist of web debate arguments, the difference in IAA is high, which might be attributed to different complexities of the web debates data. While \textsc{TvsP} only look at single arguments in isolation, often consisting of a single sentence only. 

One area of disagreement centered on arguments which were sarcastic, ironic, or included rhetorical questions. Consider the argument given in Figure~\ref{fig:ex1}, over which the expert annotators expressed disagreement. 
This argument appears to support the stance that a government has the right to censor speech, 
\cntodo{Can cut most of this graf if needed} but several linguistic cues indicate that the argument might be ironic: (a) Punctuation: ellipsis indicates thinking/searching for justifications; similarly, (b) the filler \textit{um}; (c) capitalization: the noun phrase \textit{Our Leader} is capitalized, indicating hyperbolic apotheosis; and finally, (d) the phrase \textit{(...) so I have to argue for this side.} acts like an apologia, which is put in front of the actual argument. Annotators~$1$ and ~$2$ based their judgments on an interpretation of this text that related to the estimated degree of irony in the post. While Annotator~$1$ did not perceive irony and judged the argument as \textit{very weak} in \textit{Effectiveness}, Annotator~$2$ considered it to be highly effective as in their view, the irony positively underlined the perceived stance. Annotator~$3$ gave medium scores across the board. Such disagreements were regularly discussed and usually revealed that multiple opinions may exist according to how the texts were interpreted, highlighting the high subjectivity of the task. 

Another area of disagreement was how to judge arguments on topics that were deemed ``less worthy'' of being discussed, and which usually were humorous in nature or had trivial consequences, such as \textit{Batman vs Superman}, 
in which users\cntodo{throughout: should sub out ``users'' forum participants? - LN} argued for the the superiority of either superhero.
In pilots, some experts provided lower ratings of arguments on a topic that they considered less worthy
Others thought that writing a strong, serious argument on a less worthy topic was especially difficult, and thus provided higher ratings for such arguments. 


%
%
%

\subsection{Analysis of Scores}\label{sec:analysis-scores}
\begin{figure}
     \centering
     \begin{subfigure}[t]{0.46\textwidth}
         \centering
         \includegraphics[width=1.0\linewidth,trim=0.0cm 0cm 1.5cm 0cm]{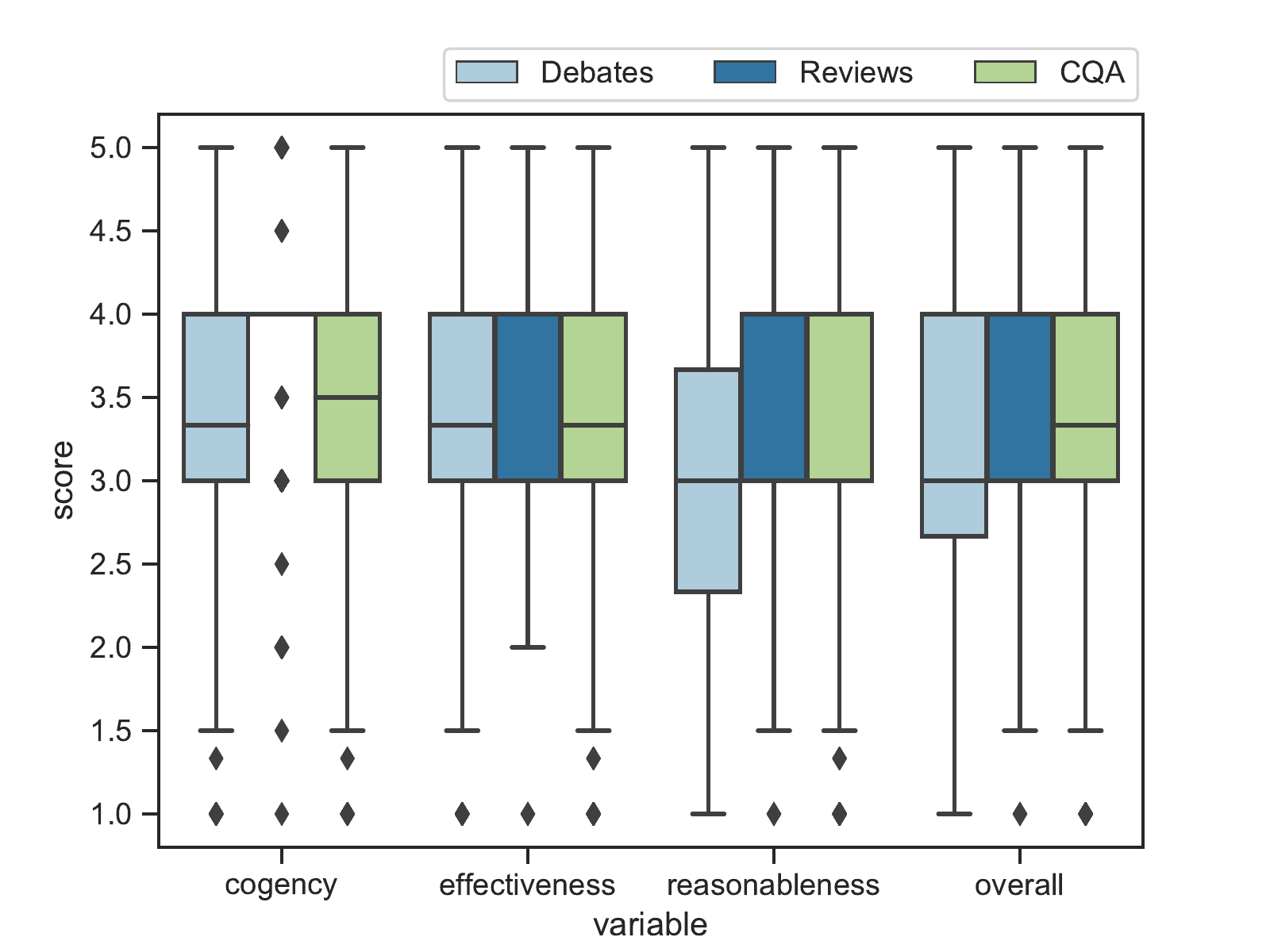}
         \caption{Expert annotations.}
         \label{fig:domains_experts}
     \end{subfigure}
     \hfill
     \begin{subfigure}[t]{0.46\textwidth}
         \centering
         \includegraphics[width=1.0\linewidth,trim=0.0cm 0.0cm 1.5cm 0cm]{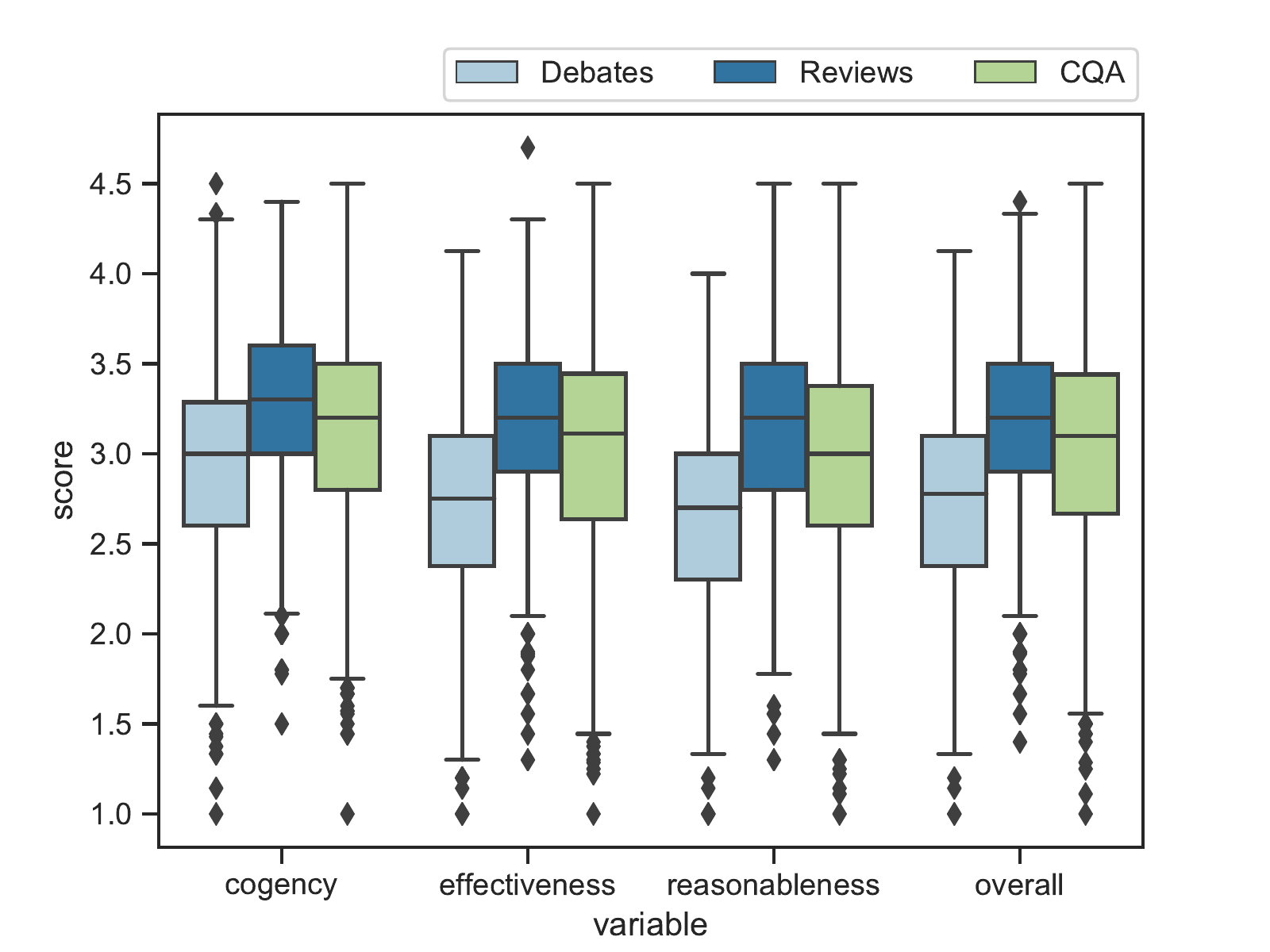}
         \caption{Crowd annotations.}
         \label{fig:domains_crowd}
     \end{subfigure}
        \caption{Score distributions 
        by domain 
        for expert and crowd annotators.}
        \label{fig:domains}
\end{figure}
%
%
%
\begin{figure*}[t]
     \centering
     \begin{subfigure}[t]{0.32\textwidth}
         \centering
         \includegraphics[width=1.01\linewidth,trim=0.0cm 0cm 1.5cm 0cm]{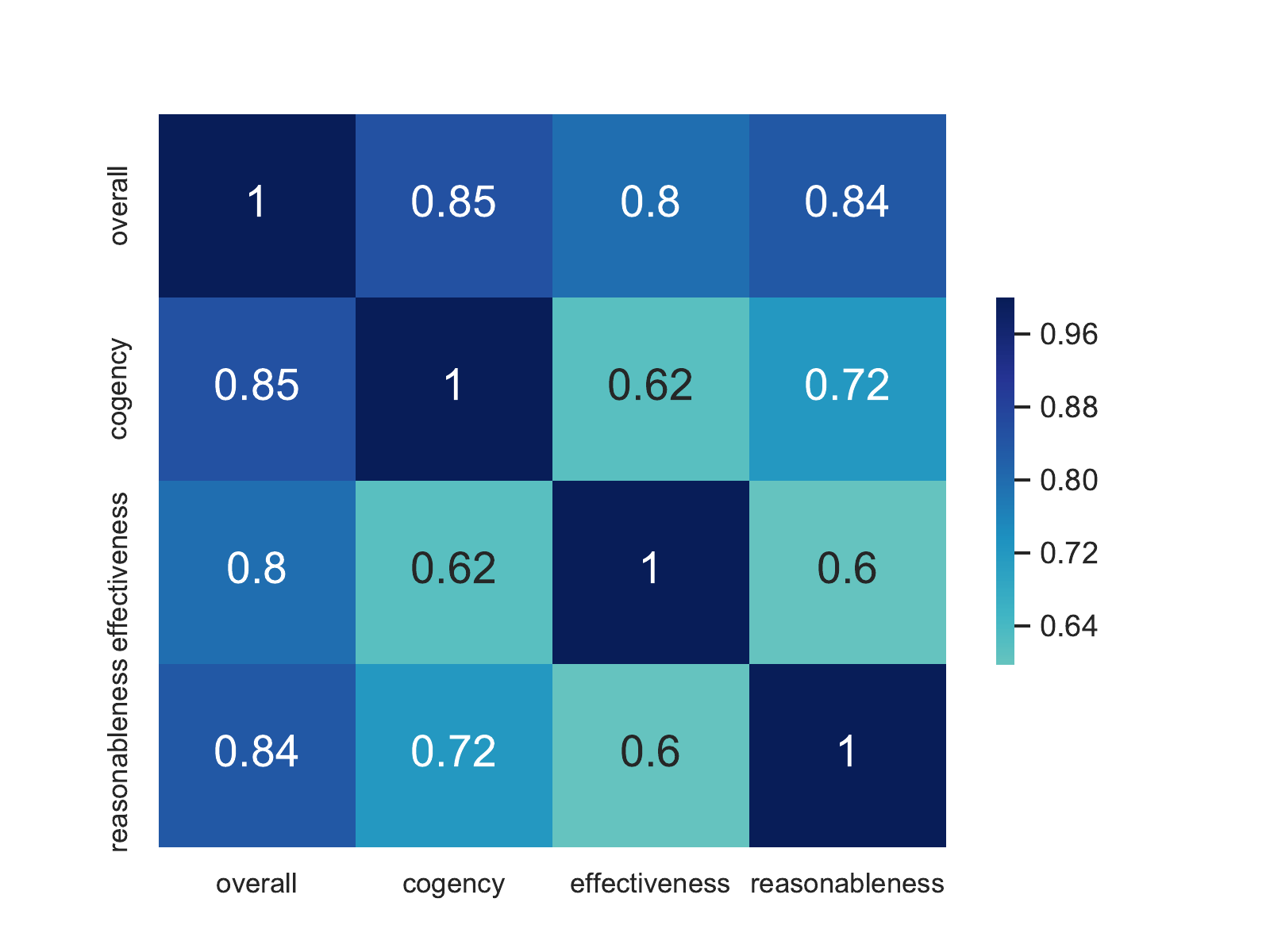}
         \caption{Experts on Debates.}
         \label{fig:corr_deb_experts}
     \end{subfigure}
    \begin{subfigure}[t]{0.32\textwidth}
         \centering
         \includegraphics[width=1.01\linewidth,trim=0.0cm 0cm 1.5cm 0cm]{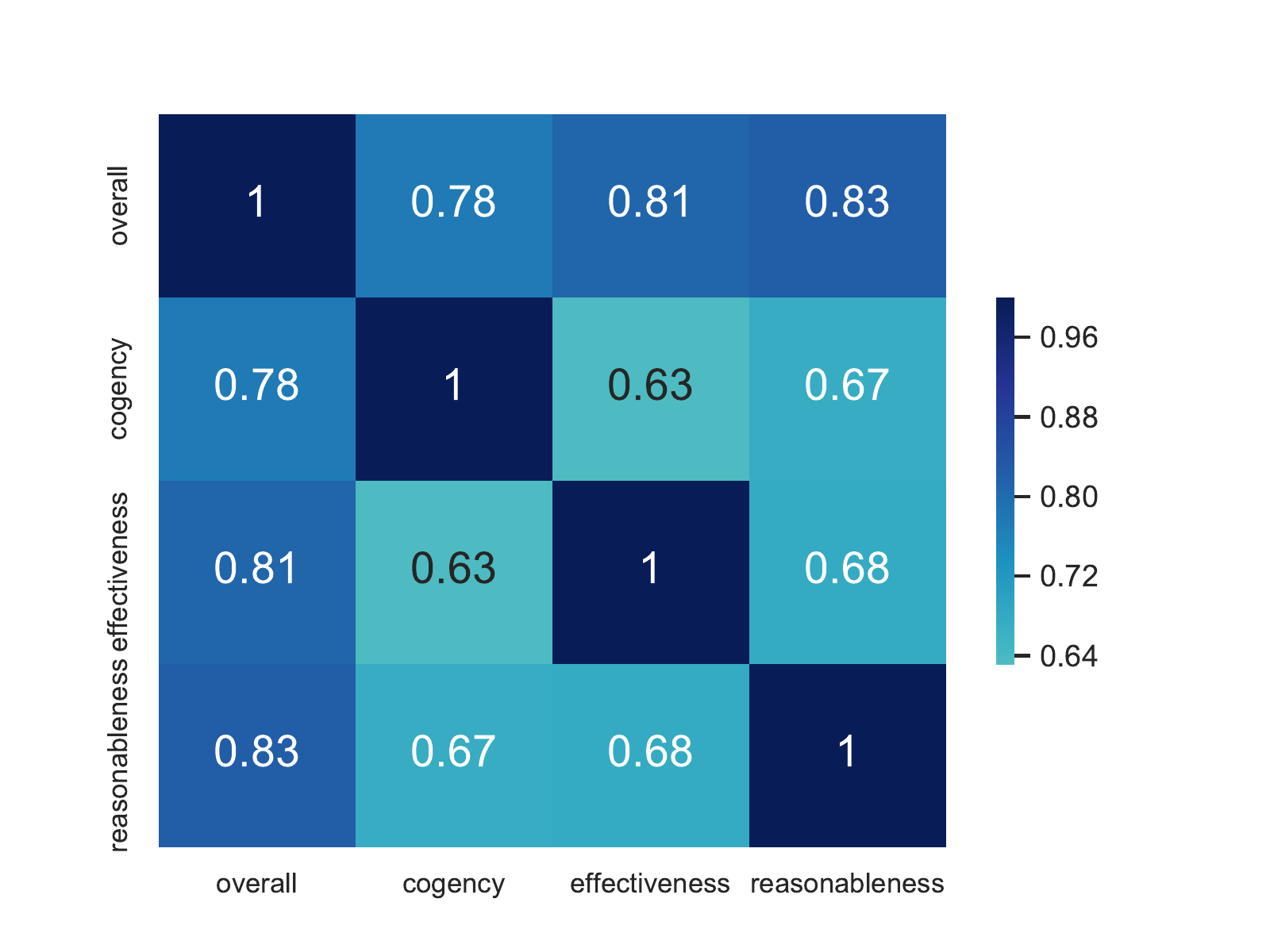}
         \caption{Experts on Reviews.}
         \label{fig:corr_rev_experts}
     \end{subfigure}
         \begin{subfigure}[t]{0.32\textwidth}
         \centering
         \includegraphics[width=1.01\linewidth,trim=0.0cm 0cm 1.5cm 0cm]{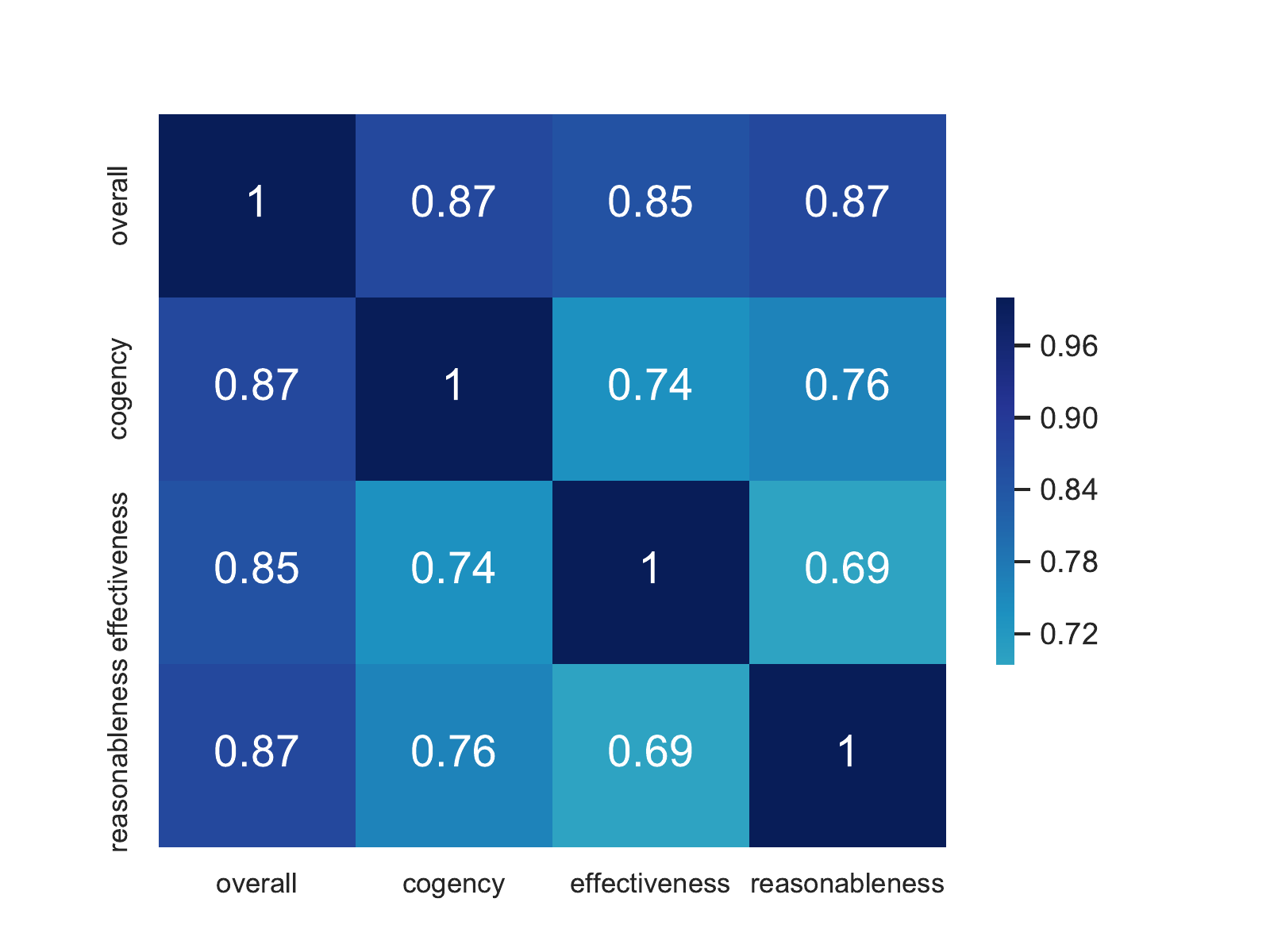}
         \caption{Experts on CQA}
         \label{fig:corr_qa_experts}
     \end{subfigure}
          \begin{subfigure}[t]{0.32\textwidth}
         \centering
         \includegraphics[width=1.01\linewidth,trim=0.0cm 0cm 1.5cm 0cm]{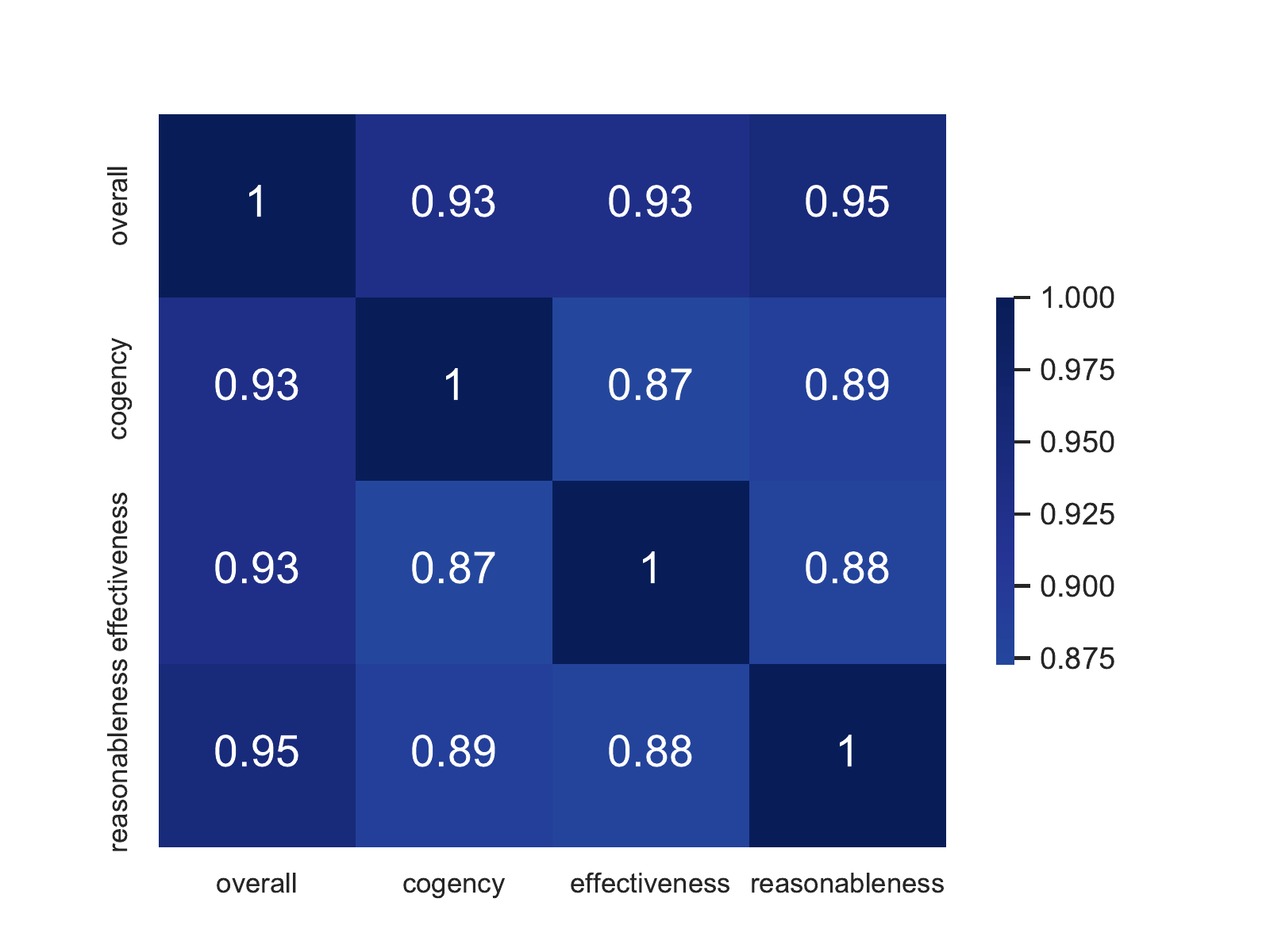}
         \caption{Crowd on Debates.}
         \label{fig:corr_deb_crowd}
     \end{subfigure}
    \begin{subfigure}[t]{0.32\textwidth}
         \centering
         \includegraphics[width=1.01\linewidth,trim=0.0cm 0cm 1.5cm 0cm]{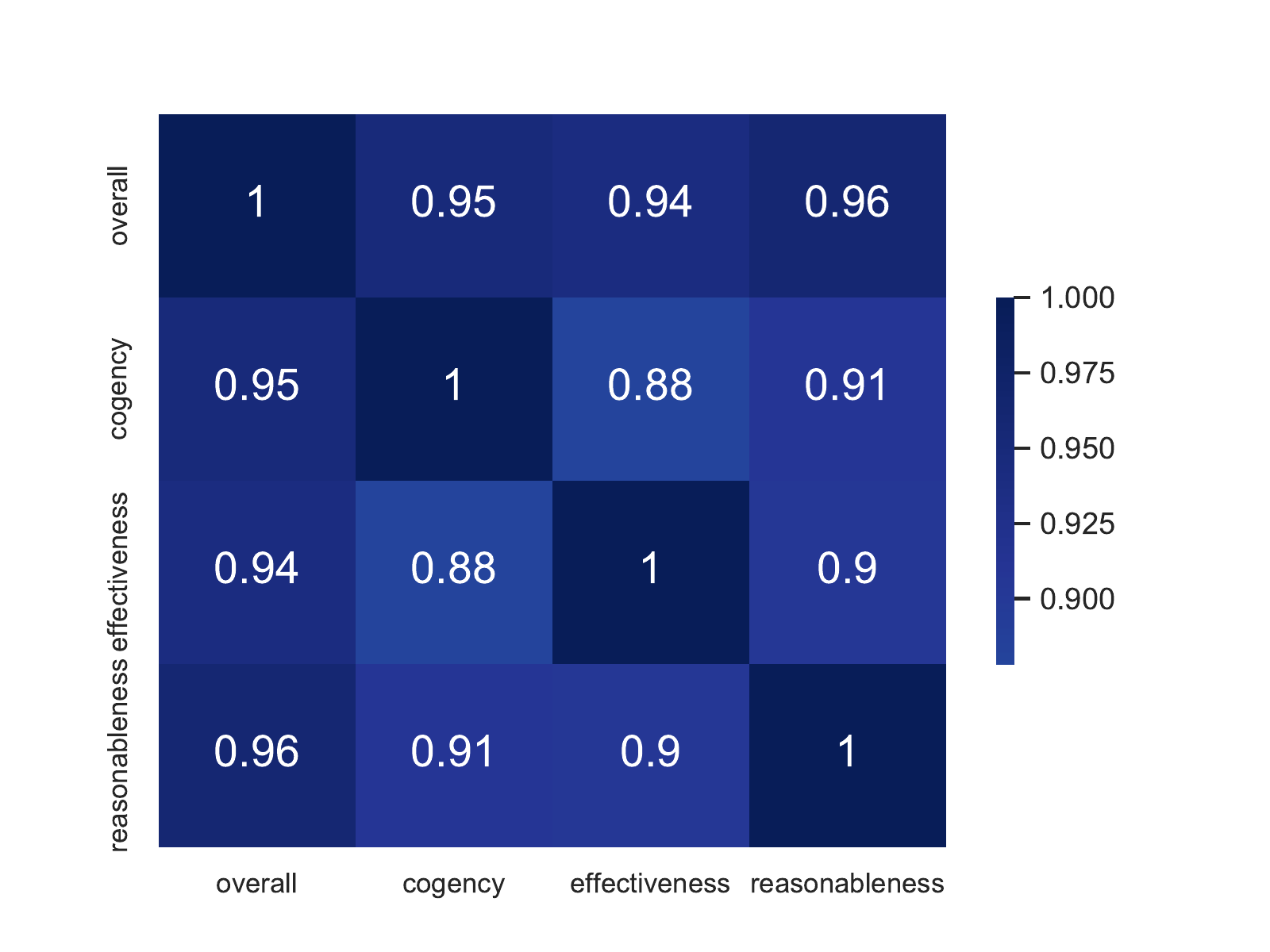}
         \caption{Crowd on Reviews.}
         \label{fig:corr_rev_crowd}
     \end{subfigure}
         \begin{subfigure}[t]{0.32\textwidth}
         \centering
         \includegraphics[width=1.01\linewidth,trim=0.0cm 0cm 1.5cm 0cm]{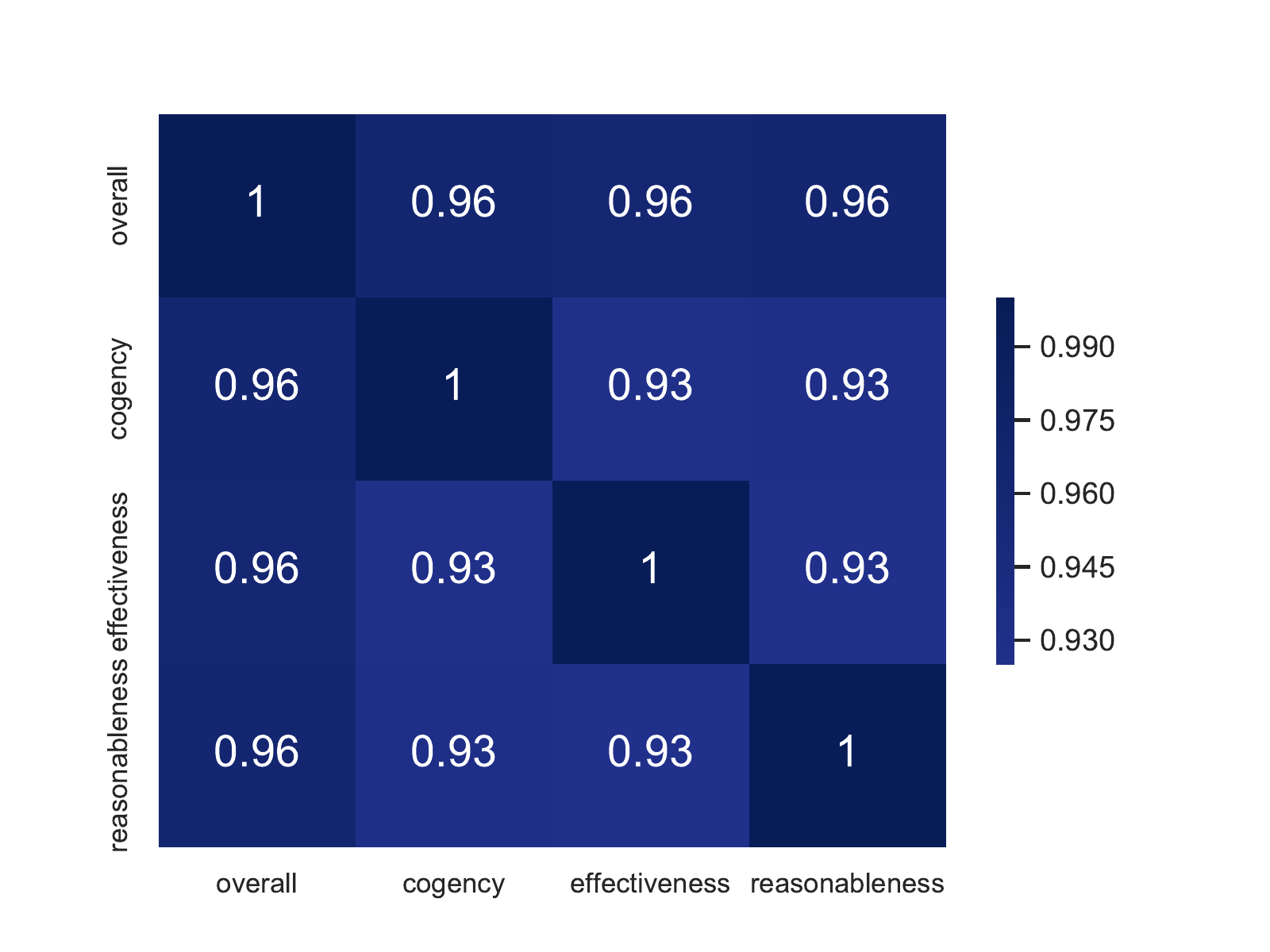}
         \caption{Crowd on CQA.}
         \label{fig:corr_qa_crowd}
     \end{subfigure}
        \caption{Mean score correlations between the different dimensions for expert and crowd annotators across the three domains (Pearson's $r$).}
        \label{fig:corrs}
\end{figure*}
The distributions of mean scores across domains and annotator groups in \corpus are depicted in Figures~\ref{fig:domains_experts} and~\ref{fig:domains_crowd}. In general, the interquartile range of the expert scores was higher than the crowd, suggesting that experts were more specific when scoring items, which is also reflected in the medians: while the crowd exhibits a tendency to score variables equally, expert annotations exhibit more differentiation.  

To understand the interrelations between Overall AQ and the dimensions, we compute Pearson correlations between the mean scores (Figure~\ref{fig:corrs}). 
Generally, the trends are similar across all three domains. For instance, for Debates (Figures~\ref{fig:corr_deb_crowd} and~\ref{fig:corr_deb_experts}), the crowd annotations exhibit stronger correlations between the different dimension  scores than the experts, with $0.83\leq r \leq 0.96$. 
Interestingly, the variance among the Pearson scores is lower, indicating that the crowd tends to distribute ratings for a single instance more consistently while the experts seem to put more weight on differentiating the dimensions. 

Expert ratings of Overall AQ have substantially stronger correlation with the dimensions than any of the dimension scores with each other, further indicating that experts are more discerning in their scores than the crowd. Across both annotator groups and all domains, the correlation between Overall AQ and Reasonableness is highest, which is consistent with earlier observations \cite{wachsmuth-etal-2017-computational}. 

%% file: 5.2-qualitative-analysis.tex
\begin{table}[t]
    \centering
    {\footnotesize
    \begin{tabular}{lp{13cm}}
    \toprule
    \textbf{Debates}\\
    \midrule
         \begin{tabular}{lc}Cogency &  2.0\\
         Effectiveness & 1.7\\
         Reasonableness  & 1.0\\
         Overall & 1.3
    \end{tabular}
         & \vspace{-.8cm}\textbf{Title:} Should you need to pass an IQ test to have kids?-- \textbf{Stance:} Dumb parents lead to more dumb kids.
\textbf{Text:} I have a strong opinion that before having children, the prospective parents should have to pass a series of background and IQ tests.
Kids being brought into this world need a good foundation to start a successful life with. You may have that limited case where the parents are morons and the kids strive to be different then their failure parents, but in most cases it is an endless line of parasites on our world.
We need more smart people.\\
\midrule
\textbf{CQA}\\
\midrule
         \begin{tabular}{lc}Cogency &  2.7\\
         Effectiveness & 2.0\\
         Reasonableness  & 1.7\\
         Overall & 2.0
    \end{tabular}
& \vspace{-.8cm}\textbf{Question:} Bounced CHECK?	
\textbf{Context:} Does the company holding the bounced check have to send you a certified letter before issuing a warrant for your arrest.  I feel almost certain that they do but i am not sure.
\textbf{Answer:} I always make sure my checks are not printed on rubber.  they are just too expensive and not worth it.  We all make a mistake from time to time, and usually it is no big deal except for the extreme annoyance and all the bounced check fees. But if you are worried about an arrest warrant then I am sure you are doing this deliberately and trying to defraud the company.  You have probably sent them a couple of bad checks already  in an attempt to string them along so your guilt is probably pretty well established.  
You can hope that you do not have to share a jail cell with a gross deviate of some sort.\\
\midrule
\textbf{Reviews}\\
\midrule
         \begin{tabular}{lc}Cogency &  1.0\\
         Effectiveness & 1.0\\
         Reasonableness  & 1.0\\
         Overall & 1.0
    \end{tabular}
& \vspace{-.8cm}\textbf{Title:} Business review: 2.0 Stars. \textbf{Business name:} Cook Out. \textbf{City:} Charlotte. \textbf{Categories:} Restaurants, Desserts, Food, Fast Food, American (Traditional), Hot Dogs, Burgers
\textbf{Review:} Burgers are good but I like those other 5 guys burgers instead oh and I guess if your not from around here don't even think about going thru the drive thru it's like the biggest most unreadable confusing hurried crazy thing ever if I ever go again hell with drive thru until I've lived here for at least 5 maybe 10 years and can be a veteran drive thru person I'm walking in it's like if I mix up all the letters in this review and give you 1 minute to read it and figure it out then you gotta move on.\\
\bottomrule
    \end{tabular}}
    \caption{Low-scoring arguments from all domains}
    \label{tab:reasonableness-examples}
\end{table}

We next examine low-scoring arguments from all domains to understand how AQ is perceived differently, focusing on the \textit{Reasonableness} dimension. 
Table~\ref{tab:reasonableness-examples} shows a low-scoring argument from each domain.\cntodo{select just one rater for each review instead of the mean}
The Debate argument raises a counterargument but does not rebut it and additionally neglects to address an obvious counterargument (i.e., the many ethical implications of such a policy). 
On the other hand, the CQA and Review arguments do not raise or address any counterarguments and are not judged Reasonable for other reasons: the CQA argument jokes about the original poster's question and accuses the poster of malignant behavior, while the Review argument delves into a personal experience that does not contribute to the discussion about the quality of the business.

%% file: 6-conclusion.tex
Theory-based AQ assessment provides a holistic and targeted perspective on AQ, but its high complexity makes annotation difficult. 
In this work, 
we describe our efforts to create \corpus, a multi-domain corpus of 5,295 arguments annotated for quality along theory-based AQ dimensions.
We demonstrate that it is possible to collect complex annotations with crowdsourcing in three domains: Debate, CQA, and Review forums.
Drawing from the initial study of \newcite{wachsmuth-etal-2017-argumentation}, which suggested the general feasibility, we relied on the intuition of trained linguists to simplify the task and guidelines while preserving the theoretical basis of the task.
The agreement between experts and the crowd was higher than the agreement in earlier studies~\cite{wachsmuth-etal-2017-argumentation}, validating our approach and indicating that it is possible to collect complex ratings using a crowd.



\corpus and the findings of our annotation study will serve as a basis for future corpus development and computational model development in theory-based AQ. They are available for download from \url{https://github.com/grammarly/gaqcorpus}.

%% file: 7-appendix.tex
\cntodo{we should submit this explicitly as supplemental material as there were some DRs because authors didn't separate them}
\begin{table}[]
    \centering
    {\footnotesize
    \begin{tabular}{lp{13cm}}
    \toprule
    \textbf{Debates}\\
    \midrule
         \begin{tabular}{lc}Cogency &  4.0\\
         Effectiveness & 4.0\\
         Reasonableness  & 4.7\\
         Overall & 4.7
    \end{tabular}
         & \vspace{-.8cm} \textbf{Title:} Torture: Can the U.S. justify it?-- \textbf{Stance:} No
\textbf{Text:} The points against torture are numerous, but the most important one is IT DOESN'T WORK. All of these ``ticking time bomb'' scenarios are fantasy right out of ``24''. That just does not happen in the real world. Another point against it: aren't we supposed to be the good guys? How in the world do we expect to keep our supposed moral superiority when we debase ourselves by stooping to the level of our enemies? If someone wants to argue we need to get down to our enemies level to really be effective in the war on terror, shouldn't we have rounded up all the German-Americans in WWII and burned them in ovens? Shouldn't we have gotten down to the German level to be really effective? I don't think the U.S. can seriously dictate to other countries when it doesn't follow the dictates of its own Constitution.\\
\midrule
\textbf{Q\&A}\\
\midrule
         \begin{tabular}{lc}Cogency &  4.3\\
         Effectiveness & 4.7\\
         Reasonableness  & 4.3\\
         Overall & 4.3
    \end{tabular}
& \vspace{-.8cm} \textbf{question:} death penality?	
\textbf{context:} i belive conviction of dui is reason for death any comments?
\textbf{answer:} ``No. Any society that values the value of a human life and promotes non-violent solutions would be hypocritical to include capital punishment in their justice system.
With a fallible justice system, there is always the risk that innocent people will be executed.
Where punnishments are decided by mortals, there is a risk that the harsher sentences are handed out according to prejudice.
Where the best barristers are available to those with the most money, there is always a risk that the poor are disadvantaged.
\\
\midrule
\textbf{Reviews}\\
\midrule
         \begin{tabular}{lc}Cogency &  4.0\\
         Effectiveness & 4.0\\
         Reasonableness  & 5.0\\
         Overall & 5.0
    \end{tabular}
& \vspace{-.8cm} \textbf{Title:} Business review: 4.0 Stars
\textbf{Business name:} Pittsburgh International Airport. \textbf{City:} Pittsburgh. \textbf{Categories:} Restaurants, Airports, Fast Food, Hotels \& Travel
\textbf{Review}: Straight-forward and easy-to-navigate airport that also provides you an unusual homely feeling!
The terminal section of PIttsburgh International Airport (PIT) is composed of four concourses extending from a central zone. The airport provides a decent selection of food and shopping in the center, so it's a decent layover stop. Throughout the premise, everything's placement makes perfect sense and the restroom facilities plus water fountains are sufficient. 
If you are flying from this airport, the TSA staff here is friendly and efficient with their job, so you don't have to worry about any delay that is commonly associated with other airport checkpoints. 
Because of its history, PIT still retained its antique feel but has not failed to provide modern airport accommodations. While it is certainly not a top-notch airport, PIT is one of the few decent airports that I have been to.
\\
\bottomrule
    \end{tabular}}
    \caption{High-scoring arguments from all domains}
    \label{tab:reasonableness-examples-high}
\end{table}

\subsection*{Annotation guidelines}


\begin{figure}[t]
    \includegraphics[scale=0.6]{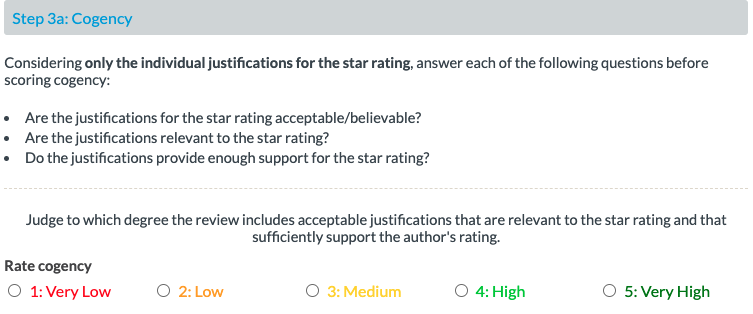}
    \centering
    \caption{Annotation Interface}
    \label{fig:interface}
\end{figure}

\subsection{Related Work}

We provide an overview on studies annotating argumentative quality as depicted in Table \ref{tbl:rw}.
\begin{table*}[!t]
\centering
{\footnotesize
\begin{tabular}{ l lllr }
\toprule
Work & Aspect & Domain 
& Annotators & \# Arguments \\ 
\midrule
\newcite{wachsmuth-etal-2017-computational} & Full taxonomy & Web debates 
&  Experts & 304 \\
\newcite{wachsmuth-etal-2017-argumentation}  & Full taxonomy & Web debates 
&  Experts \& Crowd & 304 \\
\newcite{habernal-gurevych-2016-argument} & Convincingness & Web debates 
& Crowd & 16,927 pairs\\ 
\newcite{Persing:2017:WCY:3171837.3171856} & Persuasiveness & Web debates 
& Trained annotators &  1,208\\ 
\newcite{persing2015modeling} & Argument strength & Student essays 
& Experts & 1,000 \\ 
\newcite{persing-ng-2013-clarity} &  Clarity & Student essays 
& Experts & 830 \\
\newcite{persing-ng-2014-prompt} & Prompt Adherence & Student essays 
& Experts &  830\\
\newcite{Persingorganization}  & Organization & Student essays 
& Experts & 1003\\ 
\newcite{stab-gurevych-2017-recognizing} & Sufficiency & Student essays 
& Experts? &  1,029\\ 
\newcite{wachsmuth-etal-2017-pagerank} & Relevance & Diverse genres 
& Experts & 110\\
\newcite{el-baff-etal-2018-challenge} & Convincingness & News editorials 
& Crowd Experts & 1,000 \\
\newcite{toledo2019automatic} & Overall AQ & Short Arguments 
& Crowd & 6,300\\
\newcite{gretz2019large} & Overall AQ & Short Arguments 
& Crowd & 30,000\\
\midrule
This work & Simplified taxonomy & Debates, Q\&A, Reviews 
& Experts \& Crowd & 5,285 \\
\bottomrule
\end{tabular}
}
\caption{Previous studies on annotating argument quality.
}
\label{tbl:rw}
\end{table*}

\subsection{Experimenting with different ways of including subaspect information}

The design in which annotators are presented with questions but not asked to score them is advantageous in that it is the least complex of the three.

Removing the need to score the AQ subaspects reduces task complexity but carries the risk that the subdimension definitions would be open for more subjective interpretation. While our
experts 
completed extensive training in order to ensure that each understood the full taxonomy, a challenge is how to ensure that crowd-annotations are high-quality given that the crowd cannot undergo extensive training nor be counted on to remember detailed guidelines.

We approach this challenge by presenting the subdimensions as questions to be  answered when scoring high-level dimensions. We hypothesize that this compels annotators to consider the quality of each subdimension, effectively priming them to consider the quality of each high-level dimension as a function of its subdimensions. 
We experimented with three designs with different scales for scoring the questions: (1) no scale, questions are just presented; (2) binary scale (yes/no), and (3) 3-point Likert scale.
For all of these, experts used the reduced taxonomy. 
Feedback from our experts suggested that the subdimension questions were helpful when scoring high-level dimensions, although the pilot results shown in Table \ref{tab:pilot_subdimensions} suggest that agreement was not significantly affected by the different designs.

\begin{table}
\centering
{\footnotesize
\begin{tabular}{lccccc}
\toprule
Scale & \# Arguments & Cogency & Effectiveness & Reasonableness & Overall\\
\midrule
None, just presented & 150 & 0.223 & 0.077 & \textbf{0.254} & 0.232\\
Yes/no & 150 & 0.219 & \textbf{0.331} & 0.202 & \textbf{0.328}\\
3-point Likert & 196 & \textbf{0.259} & 0.251 & 0.156 & 0.236\\
\bottomrule
\end{tabular}}
\caption{\label{tab:pilot_subdimensions} Agreement (Krippendorff's $\alpha$) between expert linguists on pilot studies with different scales for answering subdimension questions.}
\end{table}
The design in which annotators are presented with questions but not asked to score them is advantageous in that it is the least complex of the three.
To keep the task simple for both experts and the crowd while still priming annotators to consider the subdimensions in scoring, we used the first design for large-scale annotation.

\subsection*{Examples}
\paragraph{Debates.}
\begin{figure}[h]
    \centering
\begin{minipage}[t]{0.44\linewidth}\strut\vspace*{-\baselineskip}\newline
    \centering
    \includegraphics[scale=0.6]{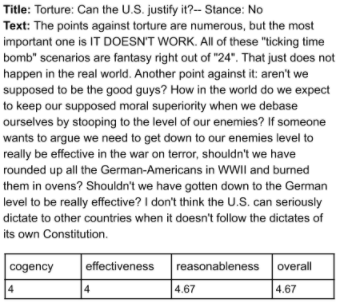}
    \caption[b]{Mean expert scores on a high-quality Debates example}
    \label{fig:debates_high_quality}
\end{minipage}
\begin{minipage}[t]{0.54\linewidth}\strut\vspace*{-\baselineskip}\newline
    \centering
    \includegraphics[scale=0.6]{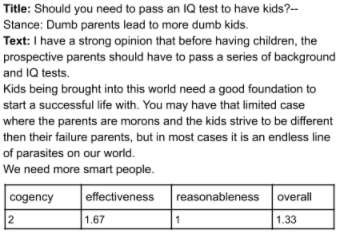}
    \caption{Mean expert scores on a low-quality Debates example}
    \label{fig:debates_low_quality}
    \end{minipage}
\end{figure}
\jttodo{figures 6 and 7 aren't very crisp.  I think we need to convert them into latex}

\paragraph{Q\&A.}
\begin{figure}[h]
    \centering
    \begin{minipage}[t]{0.54\linewidth}\strut\vspace*{-\baselineskip}\newline
        \includegraphics[scale=0.7]{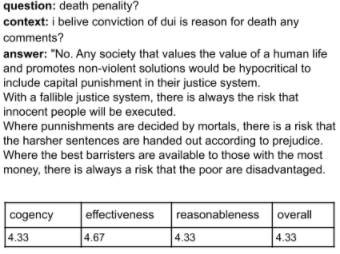}
        \caption{Mean expert scores on a high-quality Q\&A example}
        \label{fig:yahoo_high_quality}
    \end{minipage}
    \begin{minipage}[t]{0.44\linewidth}\strut\vspace*{-\baselineskip}\newline
        \includegraphics[scale=0.7]{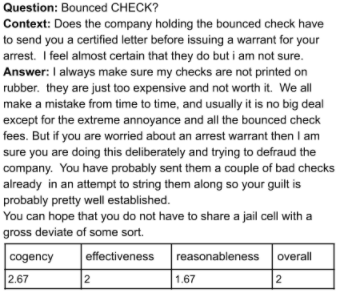}
        \caption{Mean expert scores on a low-quality Q\&A example}
        \label{fig:yahoo_low_quality}
    \end{minipage}
\end{figure}

\begin{figure}[h]
    \centering
    \begin{minipage}[t]{0.54\linewidth}\strut\vspace*{-\baselineskip}\newline
        \includegraphics[scale=0.7]{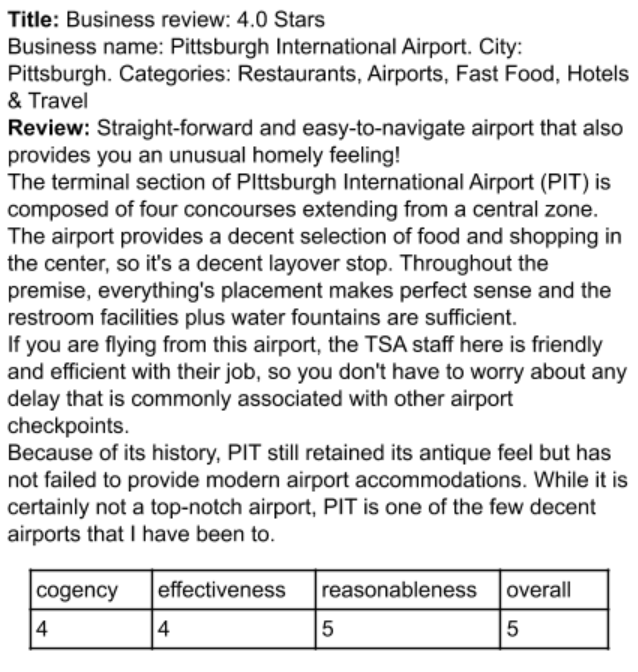}
        \caption{Expert scores on a high-quality Reviews example}
        \label{fig:yelp_high_quality}
    \end{minipage}
    \begin{minipage}[t]{0.44\linewidth}\strut\vspace*{-\baselineskip}\newline
        \includegraphics[scale=0.7]{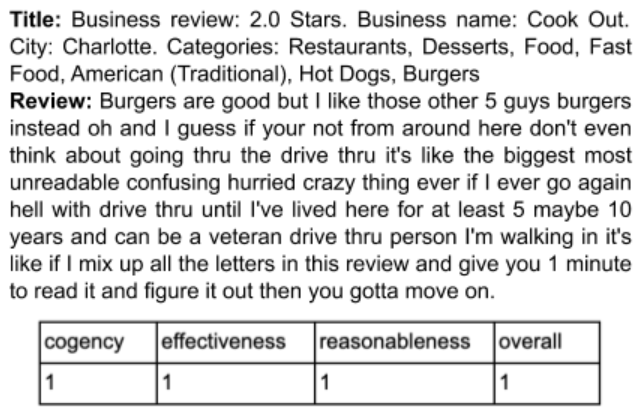}
        \caption{Expert scores on a low-quality Reviews example}
        \label{fig:yelp_low_quality}
    \end{minipage}
\end{figure}

\begin{figure}
     \centering
     \begin{subfigure}[t]{0.32\textwidth}
         \centering
         \includegraphics[width=1.0\linewidth,trim=0.0cm 0cm 1.5cm 0cm]{img/corr_debates_experts.pdf}
         \caption{Experts on Debates.}
         \label{fig:corr_deb_experts}
     \end{subfigure}
    \begin{subfigure}[t]{0.32\textwidth}
         \centering
         \includegraphics[width=1.0\linewidth,trim=0.0cm 0cm 1.5cm 0cm]{img/corr_reviews_experts.pdf}
         \caption{Experts on Reviews.}
         \label{fig:corr_rev_experts}
     \end{subfigure}
         \begin{subfigure}[t]{0.32\textwidth}
         \centering
         \includegraphics[width=1.0\linewidth,trim=0.0cm 0cm 1.5cm 0cm]{img/corr_qa_experts.pdf}
         \caption{Experts on QA}
         \label{fig:corr_qa_experts}
     \end{subfigure}
          \begin{subfigure}[t]{0.32\textwidth}
         \centering
         \includegraphics[width=1.0\linewidth,trim=0.0cm 0cm 1.5cm 0cm]{img/corr_debates_crowd.pdf}
         \caption{Crowd on Debates.}
         \label{fig:corr_deb_crowd}
     \end{subfigure}
    \begin{subfigure}[t]{0.32\textwidth}
         \centering
         \includegraphics[width=1.0\linewidth,trim=0.0cm 0cm 1.5cm 0cm]{img/corr_reviews_crowd.pdf}
         \caption{Crowd on Reviews.}
         \label{fig:corr_rev_crowd}
     \end{subfigure}
         \begin{subfigure}[t]{0.32\textwidth}
         \centering
         \includegraphics[width=1.0\linewidth,trim=0.0cm 0cm 1.5cm 0cm]{img/corr_qa_crowd.pdf}
         \caption{Crowd on QA.}
         \label{fig:corr_qa_crowd}
     \end{subfigure}
        \caption{Mean score correlations between the different dimensions for experts and crowd annotators (Pearson's $r$).}
        \label{fig:corrs-full}
\end{figure} 